\RequirePackage{fix-cm}
\documentclass[smallextended]{svjour3}       
\smartqed  
\usepackage{graphicx}
\usepackage{url}
\begin{document}

\title{Scene Text Detection for Augmented Reality - Character Bigram Approach to reduce False Positive Rate}


\author{Sagar Gubbi and Bharadwaj Amrutur
}


\institute{Indian Institute of Science}

\date{April 2017}

\maketitle

\begin{abstract}
Natural scene text detection is an important aspect of scene understanding and could be a useful tool in building engaging augmented reality applications. In this work, we address the problem of false positives in text spotting. We propose improving the performace of sliding window text spotters by looking for character pairs (bigrams) rather than single characters. An efficient convolutional neural network is designed and trained to detect bigrams. The proposed detector reduces false positive rate by 28.16\% on the ICDAR 2015 dataset. We demonstrate that detecting bigrams is a computationally inexpensive way to improve sliding window text spotters.

\end{abstract}

\section{Introduction}
\label{intro}
Augmented reality involves supplementing the view of the physical world with computer generated content. In the recent past, augmented reality (AR) smartphone apps such as Pokemon Go have garnered widespread interest. Meanwhile, high-fidelity augmented reality headsets such as Microsoft’s HoloLens and MagicLeap have been announced. These advances in AR have been made possible by recent advances in display technology, mobile platforms, and computer vision.

\begin{figure*}
  \includegraphics[width=0.95\textwidth]{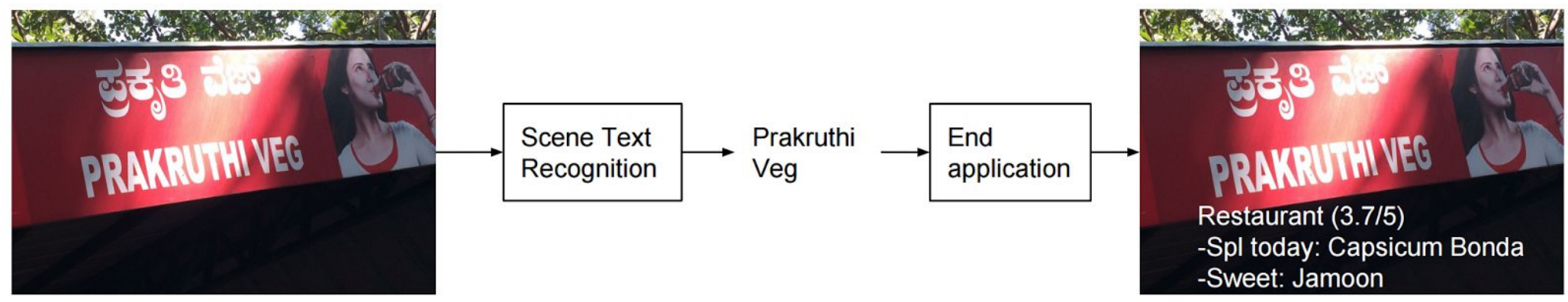}
\caption{Sample augmented reality app (fictitious) that annotates billboards with additional information using natural scene text recognition}
\label{fig:prakruthi}
\end{figure*}

Current augmented reality apps make limited use of the various sensors that are widely available. For instance, Pokemon Go mostly uses the GPS and compass. Since text is abundant in our environment, scene text recognition, i.e. extracting the text from natural images, could provide a rich source of information for building AR apps. A sample application concept that annotates billboards with additional information is shown in Fig.~\ref{fig:prakruthi}. Similarly, scene text recognition can be used to recognize and annotate books, provide information about bus routes, etc. Although scene text recognition is already used in a few applications such as Google Translate, the accuracy of text recognition on a mobile device is still not sufficient\cite{matas}\cite{jaderberg} to create a satisfying user experience.

In this work, a sliding window detector is used to find characters in natural images. A convolutional neural network, the modern tool of choice for image classification\cite{alex}, is used to detect text in each window. We propose looking for character pairs (bigrams) in each window. Our primary contribution is to show that false positive rate in detecting text can be significantly reduced by looking for bigrams rather than single characters. We also propose a neural network architecture that can efficiently detect bigrams and evaluate its performance.

The rest of this paper is organized as follows. Section~\ref{related_work} briefly goes over other works in the area of scene text recognition. The proposed text detection method is described in section~\ref{proposal}. In section~\ref{results}, experimental results are presented. Finally, section~\ref{conclusion} concludes this paper.

\section{Related Work}
\label{related_work}

Scene text recognition techniques can be broadly categorized into bottom-up detectors and top-down detectors. Bottom-up detectors work by looking for characters in the natural scene, and then cluster the detected characters into words and lines. One approach to detect characters is to filter maximally stable extremal regions\cite{matas} and reject regions that are not characters based on some features. This method is based on the observation that each character in the natural image can be segmented by binarizing the image with an appropriate local threshold. Rather than choosing a particular threshold, segmentations corresponding to all possible thresholds (0-255 for 8-bit images) are efficiently considered, and segmentations not corresponding to text are quickly filtered out with a classifier cascade. Another approach to bottom-up text detection is using a sliding window detector using a convolutional neural network to detect characters\cite{jaderberg}. Here, the classifier can observe the image through a small window (say 32$\times$32 pixels) and must decide whether a character is present in the window.

In contrast to bottom-up detectors, top-down detectors directly find words or lines of text in the image. In \cite{synthetic}, a fully convolutional network is used to locate words in an image in a manner similar to fully convolutional object detectors\cite{yolo}. This approach is refined in \cite{rnn} using a recurrent neural network to better detect word boundaries. Although top-down detectors have a significantly higher detection accuracy thanks to the benefit of additional context during detection, this comes with a substantial computational cost.

The proposed approach of looking for bigrams with a sliding window classifier is a bottom-up detector, and is thus fast compared to top-down detectors. But, by looking for bigrams rather than single characters, it retains some of the advantages of the additional context available to top-down detectors.

\section{Text spotting with convolutional neural network}
\label{proposal}

\begin{figure*}
  \includegraphics[width=0.9\textwidth]{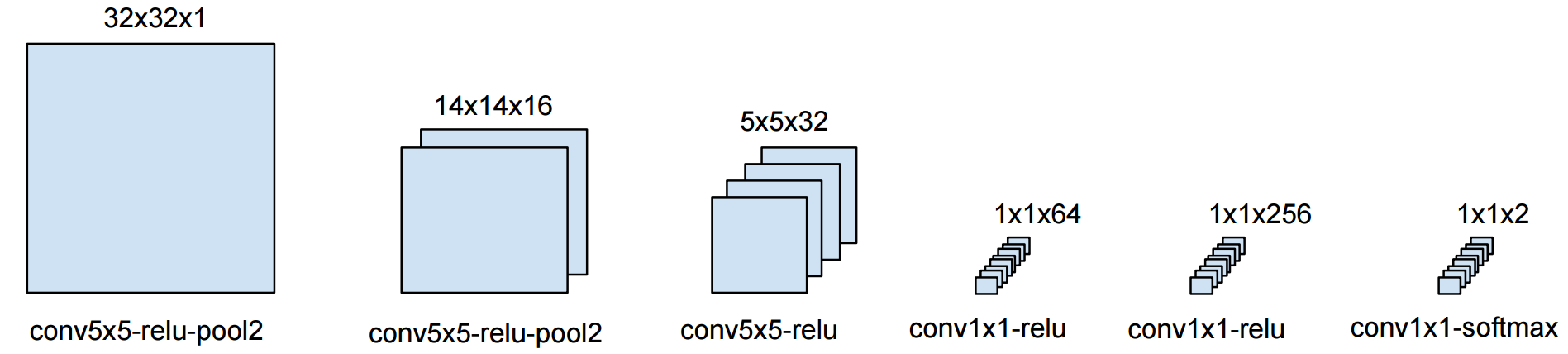}
\caption{Convolutional neural network for single character classification. The network looks at a 32$\times$32 pixel window and classifies whether a character is present in the window.}
\label{fig:conv1}
\end{figure*}

\begin{figure*}
  \includegraphics[width=0.9\textwidth]{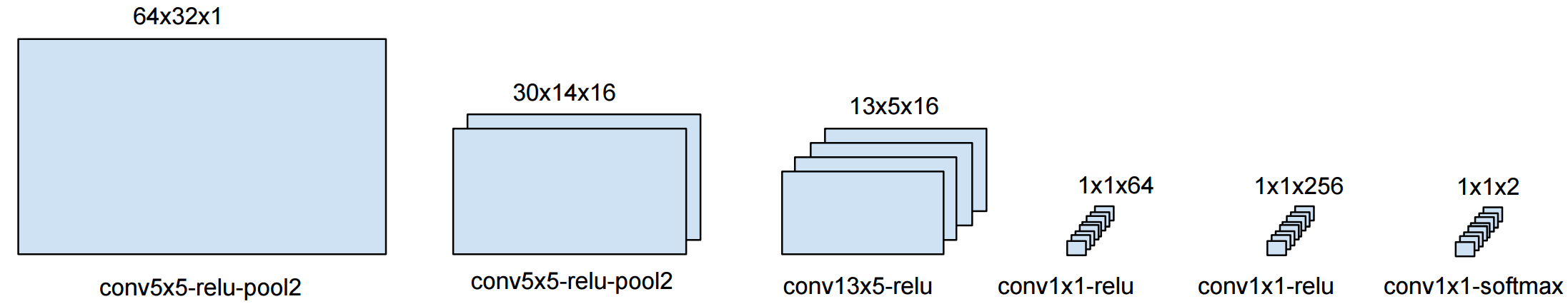}
\caption{Convolutional neural network for classifying character pairs or bigrams. The network looks at a 64$\times$32 pixel window and classifies whether a bigram is present in the window.}
\label{fig:conv2}
\end{figure*}

The sliding window text spotter works by sliding a text/no-text classifier over the entire image. The classifier looks at a small image patch (32x32~px) and decides whether a character is present in that window (characters larger than the window are detected by analyzing the same image at multiple scales). A convolutional neural network as shown in Fig.~\ref{fig:conv1} is used classify each image patch. It is similar to the architecture proposed in \cite{jaderberg}. The input is a greyscale image patch of size 32$\times$32, and the first layer of the network has 16 filters each of which is a filter of size 5$\times$5. The output of this convolution (28$\times$28~px) is passed through a rectified linear unit (ReLU) which serves as a non-linearity. This rectified activation map is max pooled over blocks of size 2$\times$2 (with stride 2$\times$2). i.e., the maximum activation in non-overlapping 2$\times$2 regions is selected and the rest of the activations are discarded. This results in an activation map of size 14$\times$14~px. The final layer is a softmax layer which treats the last layer activations as un-normalized log probabilities and squashes them into the range (0.0, 1.0). All convolutions are valid convolutions performed without padding the image (as is often done to prevent the image from shrinking in size by a few pixels after each convolution). This ensures that when the filters are applied to a larger image, they will behave identically to the training phase where examples of fixed size (32x32) are used\cite{jaderberg}. Note the absence of large filter sizes or fully connected layers. Because of this, most of the computation between successive windows is shared and the network works efficiently on a large image.

The primary difficulty with the single character classifier is that it has too little context to decide whether a character is present in the patch. In a natural scene with fine texture, a substantial number of image patches with fine details will appear similar to characters. This causes a high false positive rate. We propose expanding the sliding window horizontally to look for character pairs (bigrams) as shown in Fig.~\ref{fig:conv2}. Although fine details in the image might confuse a single character classifier, it is far less likely that such details appear similar to character pairs. This allows the classifier to achieve lower false positive rates.

\begin{figure*}
  \includegraphics[width=0.9\textwidth]{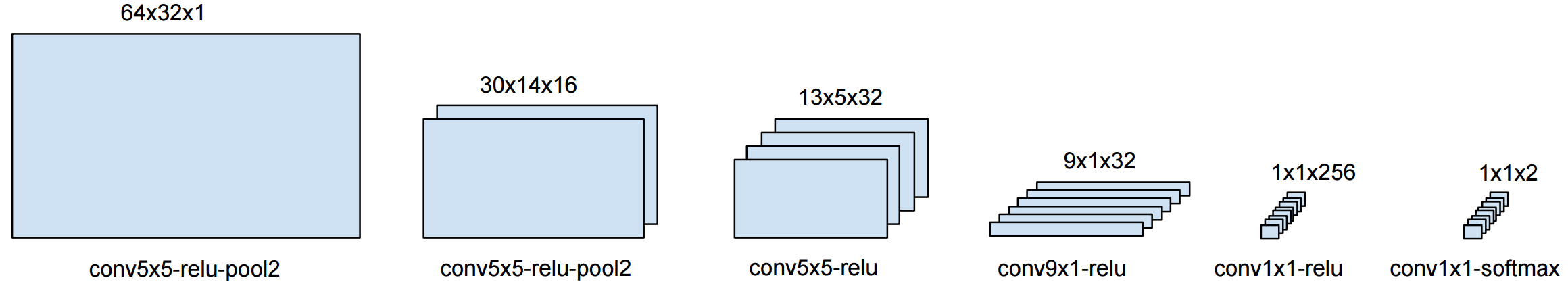}
\caption{Modified convolutional neural network for efficiently classifying character pairs or bigrams. The network looks at a 64$\times$32 pixel window and classifies whether a bigram is present in the window.}
\label{fig:conv2opt}
\end{figure*}

The expanded window size in the bigram classifier leads to increased computation. We propose mitigating this by sharing computation between letters in bigrams by adding an additional 9$\times$1 convolutional layer as shown in Fig.~\ref{fig:conv2opt}. For example, if the text ``ABC" is present in the image, the conv9x1 layer will reuse the activations around ``B" when the bigrams ``AB" and ``BC" are detected. In contrast, the unoptimized architecture in Fig.~\ref{fig:conv2} employs a large 13$\times$5 filter and would compute activations for ``AB" and ``BC" separately without any reuse.

\section{Results}
\label{results}

\begin{figure*}
  \includegraphics[width=0.75\textwidth]{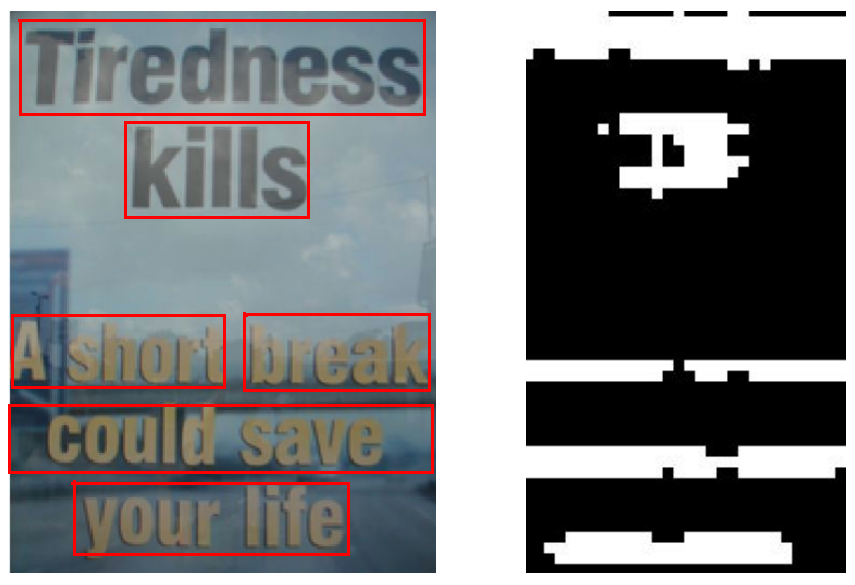}
\caption{Response of the bigram text detector on a sample image. White pixels correspond to regions positively classified by the detector as text region. The output response image was scaled with the aspect ratio fixed to improve the presentation.}
\label{fig:sample}
\end{figure*}

The dataset for training the classifier is synthetically generated\cite{synthetic} using 1,800 open fonts\footnote{\url{https://fonts.google.com/}} composited on natural images from the SUN database\cite{sun}. Linear transformations such as rotation and perspective distortion are applied to the text before compositing it with natural images with varying contrast levels. The Adam optimizer\cite{adam} with learning rate 0.001 is used with a batch size of 100. Dropout regularization (0.5) is applied in the last layer\cite{dropout}. The training is performed on an NVidia Titan X (Maxwell) GPU using Tensorflow\footnote{\url{https://www.tensorflow.org/}}.

\begin{figure*}
  \includegraphics[width=0.8\textwidth]{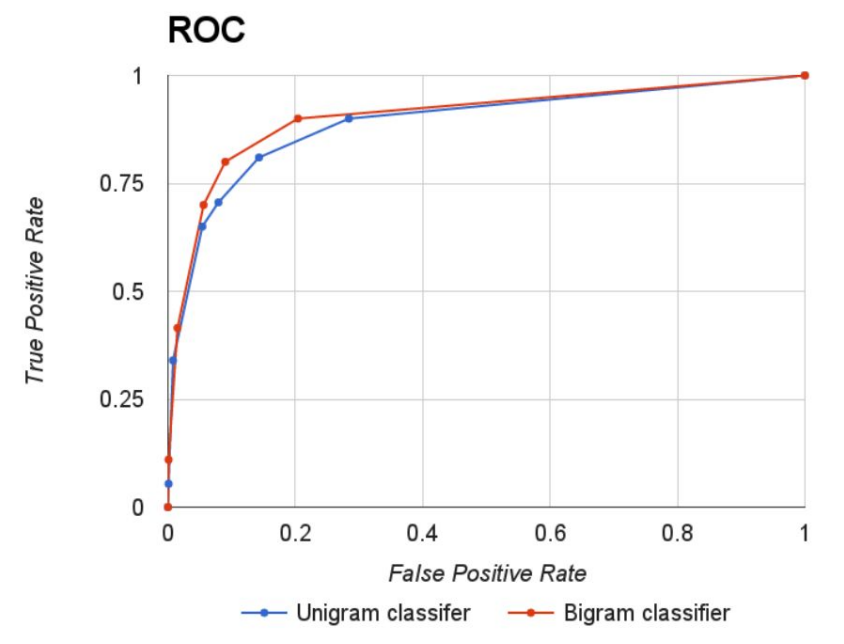}
\caption{Receiver Cperating Characteristic curve of the text classifier.}
\label{fig:roc}
\end{figure*}

A sample output of the bigram text detector is shown in Fig.~\ref{fig:sample}. Each white pixel in the output corresponds to a window where the convolutional neural network classified the patch as having text in it. Note that the output map of the detector is smaller than the input image because padding was not used for the convolutional layers. Figure~\ref{fig:roc} shows the true positive rate against the false positive rate of the classifier on the ICDAR 2015 dataset as the decision threshold (applied to the output of the softmax layer) is varied from 0.0 to 1.0. At 90\% precision, the bigram classifier reduces the false positive rate from 28.4\% to 20.4\%, a substantial reduction by 28.16\%, on the ICDAR 2015 dataset. By counting the number of operations performed by all the filters in the different detectors, we find that the optimized bigram detector in Fig.~\ref{fig:conv2opt} needs 9840 floating point multiply-accumulate (MAC) operations per pixel, which is only 25\% more than the single character classifier in Fig.~\ref{fig:conv1}. The proposed method compares favourably with other bottom-up classifiers. It achieves an f-score of 0.72 at 90\% precision compared to the f-score of 0.76 at 84\% precision in \cite{busta}. Top-down classifiers that are more computationally intensive, such as \cite{synthetic}, out-perform the proposed method with an f-score 0.84 at 93\% precision. Note that f-score is used rather than accuracy because the distribution of inputs is lopsided with most image patches having no text in them. The proposed detector operates at about 220 frames per second on 512$\times$512 px images, whereas \cite{synthetic} can process only 20 frames per second.

\section{Conclusion}
\label{conclusion}
Scene text detection could be a useful tool in building Augmented Reality applications. In such applications, a fast text spotter with high accuracy is needed to create a satisfying user experience. Bottom-up text detectors that look for characters in natural scenes suffer from high false positive rates because of the limited context available to make the classification. With a small window size, fine textures are frequently mis-classified as characters. The false positive rate can be substantially reduced by expading the window and looking for character pairs. With efficient network design, the reduction in false positive rate can be achieved with only a small computational cost.




\end{document}